\documentclass[11pt]{article}

\usepackage[final]{acl}

\usepackage{times}
\usepackage{latexsym}

\usepackage{caption}

\usepackage[T1]{fontenc}
\usepackage[utf8]{inputenc}

\usepackage{microtype}

\usepackage{inconsolata}

\usepackage{graphicx}
\usepackage{multirow} 
\usepackage{amsmath,amssymb}
\usepackage{subcaption}
\usepackage{tcolorbox}
\usepackage{makecell}

\usepackage{booktabs} 
\usepackage[table]{xcolor}

\definecolor{ForestGreen}{RGB}{34,139,34}   % muted green
\definecolor{BrickRed}{RGB}{178,34,34}     % muted red

\title{LLM Reasoning as Trajectories:\\ Step-Specific Representation Geometry and Correctness Signals}

\author{
  \textbf{Lihao Sun},
  \textbf{Hang Dong},
  \textbf{Bo Qiao},
  \textbf{Qingwei Lin},
  \textbf{Dongmei Zhang},
  \textbf{Saravan Rajmohan}
\\
\\
  Microsoft
}

\begin{document}
\raggedbottom

\pdfcompresslevel=9
\pdfobjcompresslevel=3

\maketitle
\begin{abstract}
This work characterizes large language models' chain-of-thought generation as a structured trajectory through representation space. We show that mathematical reasoning traverses functionally ordered, step-specific subspaces that become increasingly separable with layer depth. This structure already exists in base models, while reasoning training primarily accelerates convergence toward termination-related subspaces rather than introducing new representational organization. While early reasoning steps follow similar trajectories, correct and incorrect solutions diverge systematically at late stages. This late-stage divergence enables mid-reasoning prediction of final-answer correctness with ROC--AUC up to 0.87. Furthermore, we introduce trajectory-based steering, an inference-time intervention framework that enables reasoning correction and length control based on derived ideal trajectories. Together, these results establish reasoning trajectories as a geometric lens for interpreting, predicting, and controlling LLM reasoning behavior.\footnote{Code available at \url{https://github.com/slhleosun/reasoning-trajectory}.}

\end{abstract}

\section{Introduction}
Current large language models (LLMs) generate tokens by iteratively updating high-dimensional representations and decoding from them at each timestep \citep{vaswani2023attentionneed}. Given this autoregressive nature, generation can be viewed as a sequential geometric process: \textbf{a trajectory through the model's representation space}. From this perspective, when solving mathematical problems with chain-of-thought (CoT) prompting \citep{wei2023cot}, 
the sequence of reasoning steps can be viewed as successive states whose transitions collectively form a trajectory in representation space. In this work, we advance this view through a direct representational question: 
Do an LLM's reasoning steps form a structured trajectory through representation space, and if so, what does this trajectory reveal about correctness, training regimes, and opportunities for control?

Recent work has shown LLMs have representation subspaces associated with particular tasks, semantic attributes, or stages of computation \citep{li2025remaunifiedreasoningmanifold, lee2025geometryselfverificationtaskspecificreasoning, qian2025demystifyingreasoningdynamicsmutual, zhou2025geometryreasoningflowinglogics, zhou2025landscapethoughtsvisualizingreasoning, zhang2025reasoningmodelsknowtheyre, liang2025cluenonparametricverificationexperience}. Building on these insights and to operationalize the trajectory perspective, we conduct step-level analysis using explicit, naturally elicited reasoning steps (``Step 1:'', ``Step 2:'', \ldots). We extract activations immediately preceding each \texttt{Step} token, corresponding to states that reflect the prior reasoning step and precede the transition to the next. Geometrically, we find that these activations form linearly separable, step-specific subspaces. Temporally, by analyzing activation movement from one step to the next, we observe that correct and incorrect reasoning exhibit systematically different late-step dynamics. Together, this characterizes a reasoning trajectory in representation space that leads to the following observations:

\paragraph{Step-specific regions exist and organize progressively with layer depth.}
\texttt{Step}-preceding activations become increasingly separable at deeper layers. This step-specific organization is already present in the Base model, while reasoning distillation primarily reshapes this geometry by accelerating convergence toward a reasoning termination-related region at earlier layers. Moreover, the step-specific linear structure is largely shared across training regimes, indicating a common organization of representation space that is preserved despite differences in training and transfers across tasks and response formats (see Section~\ref{sec:exp1}).

\paragraph{Correct and incorrect reasoning diverge at late steps along the representation trajectory, yielding actionable correctness signals.}
By stratifying the analysis to activation movement between consecutive steps, we find that early reasoning steps follow highly similar paths for both correct and incorrect solutions, whereas late-step transitions systematically diverge. Linear classifiers trained on late-step features achieve a ROC--AUC of $0.87$ in predicting final-answer correctness prior to the emission of the final answer. Building on this signal, we further operationalize mid-reasoning, error-targeted interventions: when a predictor flags an impending failure, error-targeted test-time scaling and steering methods yield modest but consistent accuracy improvements relative to ungated counterparts and baseline (see Section~\ref{sec:exp2}).

\paragraph{Trajectory-based interventions enable correction and control of reasoning length.}
Based on correctness divergence, we introduce \emph{trajectory-based steering}, an inference-time intervention framework grounded in an \emph{ideal reasoning trajectory} derived from correct trajectories.
For correctness control, we track the model's evolving reasoning trajectory and apply low-rank steering updates when the current trajectory diverges beyond a tolerance from the ideal trajectory. This enables localized correction that nudges erroneous reasoning back toward productive computation while minimally perturbing stable, correct trajectories. For reasoning-length control, we leverage the identified termination-related subspace: steering activations toward this region accelerates convergence and shortens reasoning, whereas steering away prolongs intermediate computation (see Section~\ref{sec:exp3}).

Overall, our findings support a trajectory-based view of LLM reasoning, in which steps unfold as structured geometric motion in representation space. These trajectories contain linear signals about the phase of reasoning and the model's proximity to correct or incorrect conclusions. Leveraging these insights, we present causal evidence and actionable applications: detecting failure mid-reasoning, guiding models back toward an \emph{ideal} trajectory, and modulating reasoning length. 

\begin{figure*}[t!]
    \centering
    \includegraphics[width=\textwidth]{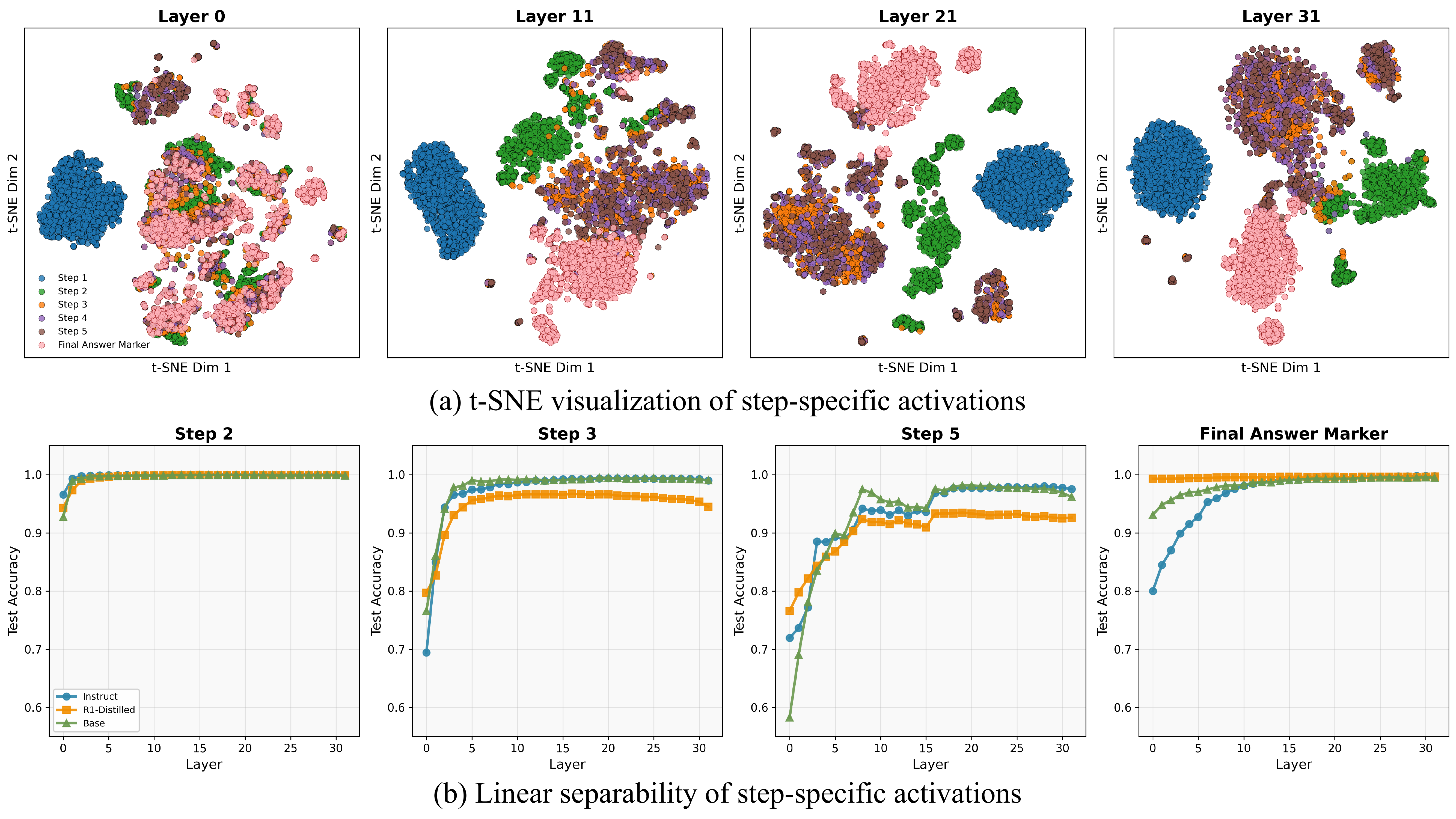}
    \caption{
    \textbf{Step-specific representation structure across layers and reasoning steps.}
    \textbf{(a)}  t-SNE visualization of activations preceding \texttt{Step} markers (Instruct model, GSM8K test split). Step-specific regions become more separated with layer depth.
    \textbf{(b)} Layer-wise linear probe accuracy for step identity prediction. $x$-axis denotes the layer from which activations are extracted, and the $y$-axis reports test accuracy of a linear classifier trained on these activations as input, with the step number as the label. Early steps are separable from shallow layers; later steps require deeper layers.
    }
    \label{fig:exp1_tsne_linearprobe}
\end{figure*}

\section{Related Work}

\paragraph{LLM CoT \& interpretability.}
Eliciting explicit step-by-step reasoning, or chain-of-thought (CoT) \citep{wei2023cot}, has become a central paradigm for LLM reasoning. 
Much work studies CoT behaviorally, examining accuracy gains, faithfulness \citep{turpin2023languagemodelsdontsay, lanham2023measuringfaithfulnesschainofthoughtreasoning, arcuschin2025chainofthoughtreasoningwildfaithful}, and behavioral limitations \citep{wang2023understandingchainofthoughtpromptingempirical,zhou2024paraphrasesolveexploringexploiting,liu2025mindstepbystep}.
Recent work applies interpretability tools to identify representation subspaces associated with specific functionalities \citep{dutta2024thinkstepbystepmechanisticunderstanding, li2025remaunifiedreasoningmanifold, lee2025geometryselfverificationtaskspecificreasoning, qian2025demystifyingreasoningdynamicsmutual, zhou2025geometryreasoningflowinglogics, zhou2025landscapethoughtsvisualizingreasoning, zhang2025reasoningmodelsknowtheyre, liang2025cluenonparametricverificationexperience,sun2026valencearousalsubspacellmscircular}, and to use internal signals for predicting model behavior \citep{zhang2025reasoningmodelsknowtheyre, liu2025llmmicroscopemodelinternals}. Among these, linear decodability is widely used as a criterion for whether information is represented in a directly accessible form for downstream computation \citep{park2024linearrepresentationhypothesisgeometry}.
In this work, we provide a step-level trajectory perspective that yields additional signals for interpreting, predicting, and controlling LLM reasoning.

\paragraph{Inference-time Intervention.}
Inference-time interventions are lightweight methods to control model behavior without retraining. 
For reasoning, such methods generally elicit additional reasoning, commonly through test-time scaling by injecting specific tokens \citep{muennighoff2025s1simpletesttimescaling, zhang2025alphaonereasoningmodelsthinking} or via activation steering \citep{turner2024steeringlanguagemodelsactivation, zhang2024bestpracticesactivationpatching}. 
Though powerful, recent work highlights their limitations when applied unconditionally \citep{ghosal2025doesthinkinghelpmirage, wang2025waitdontneedwait, zhao2025testtimescalingreasoningmodels}. 
Motivated by these limitations, we explore error-prediction gating toward more targeted interventions, and further propose trajectory-based interventions that operate adaptively on the evolving representation trajectory rather than injecting fixed tokens.

\section{Step-specific Representation Subspaces}
\label{sec:exp1}

\subsection{Datasets \& Models}
We focus on math tasks, where step-based reasoning is naturally elicited, using the GSM8K dataset (7,473 training and 1,319 test questions) and MATH-500 (500 questions) \citep{cobbe2021gsm8k, hendrycks2021math500}. For controlled experiments, we impose a fixed reasoning format via a standard zero-shot CoT prompt template (see Appendix~\ref{sec:prompt}) that requires each reasoning step to begin with an explicit marker (\texttt{Step}) and the final answer preceded by a termination marker (e.g., \texttt{\#\#\#\#}).

To evaluate effects of different training regimes, we study three variants of Llama~3.1~8B using deterministic generation: Base, the pretrained model without instruction tuning \citep{grattafiori2024llama3herdmodels}; Instruct, obtained from Base via supervised instruction tuning and preference alignment \citep{grattafiori2024llama3herdmodels}; and R1-Distill, a reasoning model built on Llama~3.1~8B via chain-of-thought distillation \citep{deepseekai2025deepseekr1incentivizingreasoningcapability}.

By holding the dataset, prompting format, and decoding strategy constant, we isolate differences arising from training regimes and internal computation, enabling a controlled comparison of reasoning subspaces and trajectory structure.

\subsection{Extracting Step-specific Activations}
For each question, we extract hidden activations at two classes of decoding timesteps:
(i) the token immediately preceding each \texttt{Step} marker (denoted $\mathbf{h}^{(\ell)}_{t(\text{Step } k)-1}$), which captures the model's internal state after completing the $k$-th reasoning step and before transitioning to the next; and (ii) the token immediately preceding the final answer marker ($\mathbf{h}^{(\ell)}_{t(\text{term})-1}$), which corresponds to when the model has completed reasoning and is ready to emit the final answer. The ordered sequence $\mathbf{h}^{(\ell)}_{t(\text{Step }1)-1},\;
\mathbf{h}^{(\ell)}_{t(\text{Step }2)-1},\;
\ldots,\;
\mathbf{h}^{(\ell)}_{t(\text{term})-1}$ thus provides snapshots of the model's representation trajectory as reasoning unfolds.

\subsection{Analyzing Step-specific Subspaces}
To characterize how step-specific activations are organized in representation space, we use t-SNE for qualitative visualization and train a binary classifier for each Step~$X$ against all other steps, to quantify the linear separability of step-specific activations, testing whether each step occupies a distinct region of representation space.

\paragraph{Step-specific and termination-related activations occupy highly linearly separable regions in representation space.} 
Quantitatively, early steps exhibit linear structure across all layers and training regimes (see Figure~\ref{fig:exp1_tsne_linearprobe}). Notably, final answer marker and early-step activations are exceptionally distinct: Step~1 has probe accuracy above $0.99$ at every layer for all models, and Step~2 reaches this accuracy level by layer~2 (Figure~\ref{fig:exp1_tsne_linearprobe}b). Steps~3,~4, and~5 can also reach near-ceiling probe accuracy for the Instruct and Base models after substantial depth, with step-specific activations moving into increasingly well-delineated regions as the model processes into deeper layers. 

Crucially, the separability is across step numbers. If the model were simply encoding "a step marker is imminent" as an ordinary sentence, all pre-step activations would cluster together regardless of step index. However, in our experiment, every step occupies distinct representation regions separate from others, indicating that the model tracks \emph{where} it is in the reasoning progress. Moreover, because we extract activations at the token immediately before each \texttt{Step} marker, these activations reflect the state upon completing the prior reasoning step---capturing accumulated computation carried forward to subsequent timesteps rather than surface formatting cues. Together with the cross-task transfer results in Section~\ref{sec:cross_task}, these findings suggest that step-specific activations are organized into linearly separable regions of representation space.

\paragraph{Step-specific activations become progressively less entangled with depth.}
As shown in Figure~\ref{fig:exp1_tsne_linearprobe}a, step-specific activations transition from heavily intermixed regions at early layers to more distinct regions at deeper layers, particularly for Steps~3,~4, and~5. Quantitatively, for Step~5, linear probe accuracy at layer~0 ranges from $0.58$ (Base) to $0.77$ (R1-Distill), and exceeds $0.90$ only after layer~6. These results indicate that increasing depth progressively disentangles step-specific activations into more stable regions corresponding to different steps, with different training regimes exhibiting distinct \textit{rates} of this organization.

\subsection{Training Regimes Reshape Step Geometry}

\begin{table}[t]
\centering
\scriptsize
\setlength{\tabcolsep}{3pt}
\begin{tabular}{llccccc}
\toprule
\textbf{Probe} &
\textbf{Eval.} &
\multicolumn{4}{c}{\textbf{Step}} &
\textbf{Final Ans.} \\
\textbf{From} &
\textbf{On} &
2 & 3 & 4 & 5 &
\textbf{Marker} \\
\midrule

Instruct
& R1-Dist. & 0.99$_{\tiny L18}$ & 0.93$_{\tiny L19}$ & 0.87$_{\tiny L12}$ & 0.91$_{\tiny L18}$ & 0.87$_{\tiny L23}$ \\
& Base     & 1.00$_{\tiny L12}$ & 0.97$_{\tiny L08}$ & 0.95$_{\tiny L08}$ & 0.93$_{\tiny L18}$ & 0.97$_{\tiny L21}$ \\
\midrule

R1-Dist.
& Instruct & 1.00$_{\tiny L03}$ & 0.92$_{\tiny L08}$ & 0.88$_{\tiny L07}$ & 0.93$_{\tiny L27}$ & 0.98$_{\tiny L19}$ \\
& Base     & 1.00$_{\tiny L06}$ & 0.94$_{\tiny L04}$ & 0.89$_{\tiny L30}$ & 0.91$_{\tiny L08}$ & 0.96$_{\tiny L19}$ \\
\midrule

Base
& Instruct & 1.00$_{\tiny L21}$ & 0.98$_{\tiny L18}$ & 0.97$_{\tiny L18}$ & 0.97$_{\tiny L23}$ & 1.00$_{\tiny L31}$ \\
& R1-Dist. & 0.99$_{\tiny L12}$ & 0.91$_{\tiny L18}$ & 0.90$_{\tiny L18}$ & 0.92$_{\tiny L17}$ & 0.94$_{\tiny L02}$ \\
\bottomrule
\end{tabular}

\caption{
\textbf{Cross-model transfer of step-specific linear probes. }
Each entry is the best accuracy across layers; subscripts indicate the peak layer. Step 1 omitted (1.00 for all pairs).
}
\label{tab:linear_probe_transfer}
\end{table}

As shown in Figure~\ref{fig:exp1_tsne_linearprobe}b, step-specific structure is \textbf{already present in the Base model, while reasoning training accentuates termination-related structure from the earliest layers}. Specifically, final answer markers achieve probe accuracy above $0.99$ at layer~0 in R1-Distill, compared to $0.80$ for Instruct and $0.93$ for Base. 
This ``faster convergence'' refers to \emph{layer-depth} convergence---i.e., R1-Distill's activations are separable into the termination subspace at shallower layers---not to the number of output reasoning steps (see Appendix~\ref{sec:step_count_dist} for step-count distributions across models). At the same time, R1-Distill exhibits consistently lower separability for Steps~3,~4, and~5. 
This pattern indicates that reasoning training does not uniformly strengthen step-specific structure. It accelerates movement into the termination subspace while leaving some intermediate steps less separable, though still concentrated within coherent regions with probe accuracy consistently above $0.90$. 

\paragraph{Step-specific regions are largely shared and preserved across training regimes at the level of linear structure.}
As shown in Table~\ref{tab:linear_probe_transfer}, cross-model transfer of step-specific linear probes consistently achieves accuracy above $0.90$ for nearly all model pairs. Notably, layer-averaged transfer accuracies also mostly exceed $0.90$ (see Appendix~\ref{sec:linear_probe}). The consistently high transfer accuracies indicate that step-specific linear structure is largely shared across training regimes. Therefore, post-training primarily reshapes the depth at which step-specific structure becomes most linearly salient, while preserving a common underlying organization.

\begin{figure*}[t]
    \centering
    \includegraphics[width=\textwidth]{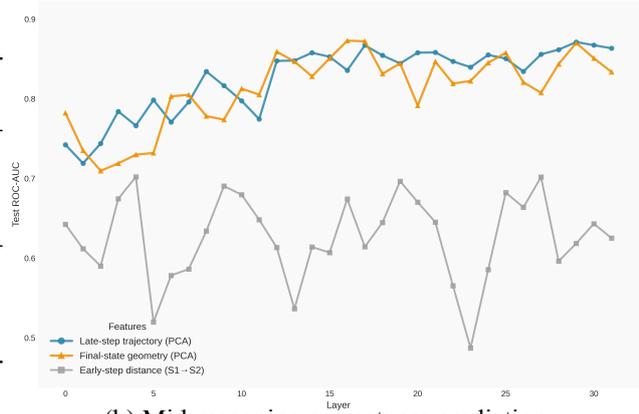}
    \caption{
    \textbf{Trajectory divergence and mid-reasoning correctness signals} (Instruct model, GSM8K).
    \textbf{(a)} Late between-step activation distances diverge between correct and incorrect trajectories, while early transitions remain similar; $^{\dagger}$ marks transitions where 95\% CIs of correct and incorrect distances do not overlap. 
    \textbf{(b)} Late-step trajectory features predict correctness with average AUC $\approx 0.83$ (peak $0.87$ at layer~29) vs. $\approx 0.63$ for early-step features. Curves are from a single random seed; see Table~\ref{tab:predictor_baselines} for seed-averaged values.
    }
    \label{fig:exp2_dist_correctness}
\end{figure*}

\subsection{Robustness to Prompt Format}
\label{sec:freeform}

The preceding analyses use a fixed-form prompt that explicitly elicits \texttt{Step} markers. To test whether the observed geometry is primarily driven by surface-level formatting, we use a minimal prompt with no formatting constraints.\footnote{\texttt{"Solve the following problem. Think step by step.\textbackslash n\textbackslash nQuestion: \{question\}\textbackslash n\textbackslash nSolution:\textbackslash n"}.}

\begin{table}[t]
\centering
\small
\setlength{\tabcolsep}{4pt}
\begin{tabular}{lrrl}
\toprule
\textbf{Format Category} & \textbf{Count} & \textbf{\%} & \textbf{Segments} \\
\midrule
\texttt{Step X:}           & 4{,}820 & 64.5 & 4.6 steps \\
\texttt{\textbackslash n\textbackslash n}-sep.\ paragraphs & 838 & 11.2 & 5.6 paragraphs \\
\texttt{\textbackslash n}-separated lines        & 687  & 9.2  & 7.6 lines \\
Single block               & 662  & 8.9  & 7.7 sentences \\
Numbered list (1./1))      & 467  & 6.2  & 4.4 items \\
\bottomrule
\end{tabular}
\caption{
\textbf{Response format distribution under minimal freeform prompting} on GSM8K (Instruct model). Without explicit formatting instructions, the model spontaneously adopts \texttt{Step X:} formatting in $64.5\%$ of responses. Segment counts report per-response means.
}
\label{tab:freeform_categories}
\end{table}

Under this setting on GSM8K, the Instruct model spontaneously produces five distinct response formats (Table~\ref{tab:freeform_categories}). Notably, the model uses \texttt{Step X:} formatting in $64.5\%$ of cases without explicit instruction, suggesting that step-marked reasoning is naturally favored rather than a prompt artifact. For the remaining responses, we extract activations at natural structural boundaries (e.g., paragraph breaks, sentence-ending punctuation, numbered markers).

We then apply linear probes trained on fixed-form activations to these freeform activations. Despite relatively low fixed-form vs.\ freeform activation similarity (mean cosine $\approx 0.41$, CKA $\approx 0.48$), we observe strong transfer: best-layer probe accuracies reach $0.93$ (Step~1), $0.84$ (Step~2), $0.83$ (Step~3), $0.84$ (Step~4), $0.88$ (Step~5), and $0.92$ (Answer). Crucially, restricting to non-\texttt{Step} format categories yields comparably strong results, with best-layer accuracies consistently above $0.84$ across all steps (see Appendix~\ref{sec:freeform_details} for full results). 

As a control, probes trained on randomly shuffled step labels achieve only $0.59 \pm 0.04$ average accuracy. 
These results indicate that the observed step-specific geometry reflects genuine reasoning progress rather than formatting artifacts. 

\section{Correctness in Trajectory Geometry}
\label{sec:exp2}

Having identified task-specific structure in representation space in Section~\ref{sec:exp1}, we next examine trajectories temporally by analyzing the paths connecting steps. Here, we group trajectories by final-answer correctness as a behavioral label.

Concretely, for two steps $a$ and $b$, we compute distances
$d\!\left(\mathbf{h}^{(L)}_{t(a)-1},\; \mathbf{h}^{(L)}_{t(b)-1}\right)$
using Euclidean distance to capture the magnitude of movement in representation space, and cosine distance to capture changes in direction. We focus on activations from the final layer, as it reflects the cumulative outcome of computation after all intermediate processing, and because step-specific subspaces are generally most clearly separated at this depth (Section~\ref{sec:exp1}).

\subsection{Trajectory Distance Difference}
\label{sec:dist_body}

As shown in Figure~\ref{fig:exp2_dist_correctness}a, \textbf{early-step geometry is mostly correctness-invariant.} Quantitatively, it exhibits no statistically significant distance difference signal regarding final-answer correctness. The $\Delta$(I--C) values for Step~1$\rightarrow$2 are near zero and fall within 95\%CI under both Euclidean and cosine metrics. This is consistent with the findings in Section~\ref{sec:exp1} that Step 1 and 2 subspaces are highly stable with near-perfect probe accuracies. Together, this indicates that the model traverses a common geometric path in representation space during initial problem processing and early-stage reasoning. 

In contrast, \textbf{late-step trajectory distance differences show correctness-related divergence.} For the transition from the second-last step to the last step, incorrect solutions exhibit statistically significant differences, with Euclidean $\Delta$(I--C)$=-4.26$ and cosine $\Delta$(I--C)$=-0.02$. This divergence becomes more pronounced in the final transition from the last step to the answer marker, where the differences increase to $-13.39$ (Euclidean) and $-0.06$ (cosine). Although correct and incorrect trajectories initially follow similar geometric paths and ultimately converge to concentrated step-specific regions, the paths taken to reach the regions become increasingly different as reasoning progresses and deviations accumulate. 
% This provides a concrete signal for mid-reasoning prediction of final-answer correctness.

\subsection{Mid-reasoning Correctness Prediction via Trajectory Divergence}
\label{sec:predictors}
Motivated by the observed late-step geometric divergence, we next use trajectory signals to predict final-answer correctness \emph{before} answer emission. Concretely, we train an $\ell_2$-regularized logistic regression classifier on activation features extracted from the Instruct model using the GSM8K training split, and evaluate its ability to distinguish incorrect from correct solutions on a held-out test split. Performance is reported in terms of test ROC--AUC.\footnote{Unless otherwise noted, results in this section are from a random fixed seed (seed 42); seed-averaged results with variance are reported in Table~\ref{tab:predictor_baselines} and Appendix~\ref{sec:exp_details}.} We consider a range of feature constructions drawn from different portions of the reasoning trajectory, including early-step geometry (Step~1 and Step~2), late-step trajectory (concatenating the final answer marker activation with the last-step transition with PCA dim=$128$), and final-state representations extracted immediately before the answer marker (PCA dim=$128$).

\paragraph{Late-step trajectory geometry predicts correctness with average AUC of 0.83 across layers, whereas early-step geometry only achieves 0.63.}
As shown in Figure~\ref{fig:exp2_dist_correctness}b, late-step trajectory features are more predictive of final-answer correctness than early-step geometry. Specifically, using late-step trajectory features, we achieve an average AUC of $0.83$ across layers, with a peak AUC of $0.87$ at layer~29. Using the final answer marker activations alone already yields strong predictive performance, with an average AUC of $0.81$. This indicates that the model's terminal representation encodes substantial information about final-answer correctness, consistent with recent findings on correctness prediction in MMLU-style settings \citep{liu2025llmmicroscopemodelinternals}.

In contrast, feature sets derived from early-step geometry perform markedly worse. Using only the Step~1$\rightarrow$2 activation difference achieves an average AUC of $0.63$, only modestly above chance. Directly concatenating the Step~1 and Step~2 activations yields a lower average AUC of $0.61$. Including an additional early step (Step~3) reaches an average AUC of $0.63$. These results mirror Figure~\ref{fig:exp2_dist_correctness}a, where early-step transitions show no statistically significant correctness-related divergence, while late-step geometry exhibits consistent separation between correct and incorrect trajectories.

Notably, the strongest predictors are not raw activations but trajectory-based \texttt{Step}-difference features. This supports the interpretation that correctness is reflected not merely in where the model ends up in representation space, but in how it gets there. Together with the distance-based results in Section~\ref{sec:dist_body}, these findings provide evidence that trajectory divergence provides a concrete mid-reasoning signal for final-answer correctness.

\begin{table}[t]
\centering
\small
\setlength{\tabcolsep}{6pt}
\begin{tabular}{lc}
\toprule
\textbf{Method} & \textbf{Best-layer AUC} \\
\midrule
Step-count only         & $0.649 \pm 0.021$ \\
LogitLens (best config) & $0.765 \pm 0.027$ \\
Trajectory features (ours) & $0.852 \pm 0.039$ \\
\bottomrule
\end{tabular}
\caption{
\textbf{Correctness predictor comparison.}
Trajectory features substantially outperform both a step-count-only baseline and logit-level features (entropy, answer-marker token rank, top-1 probability) at step boundaries. Values report best-layer AUC averaged over three seeds. 
}
\label{tab:predictor_baselines}
\end{table}

\paragraph{Trajectory features outperform logit-level and length-based baselines.}
To rule out surface-level confounds, we compare against step-count-only and logit-lens baselines (Table~\ref{tab:predictor_baselines}). A classifier using only the number of reasoning steps as a feature achieves AUC $0.649 \pm 0.021$, indicating that length carries some signal but falls well below trajectory-based prediction. Logit-lens features at step boundaries (entropy, answer-marker token rank, and top-1 probability) achieve a best AUC of $0.765 \pm 0.027$. Trajectory features ($0.852 \pm 0.039$) substantially outperform both. Under length-balanced resampling, trajectory features still achieve AUC $0.847 \pm 0.006$, only $\sim\!0.02$ below the original, confirming that the signal is not driven by length. 

\subsection{Toward Error-targeted Inference-time Interventions}
\label{sec:error_aware_intervention}

We next explore how this signal can be used to enhance inference-time intervention methods, with the goal of intervening only when an impending failure is detected. This aims to mitigate the \emph{overthinking} drawback of unconditional interventions, which can unintentionally degrade originally correct reasoning \citep{ghosal2025doesthinkinghelpmirage, wang2025waitdontneedwait, zhao2025testtimescalingreasoningmodels}. Here, we evaluate two common classes of inference-time interventions: test-time scaling and activation steering.

\begin{table}[t]
\centering
\footnotesize
\setlength{\tabcolsep}{4pt}
\begin{tabular}{lccc}
\toprule
\textbf{Intervention} & \textbf{Always} & \textbf{Gated} & \textbf{Gain vs Always} \\
\midrule
\texttt{Step}             & $-1.59$  & $+0.91$  & $+2.50$ \\
\texttt{Check}            & $-11.70$ & $+0.23$  & $+11.90$ \\
\texttt{Wait}             & $-30.50$ & $-0.68$  & $+29.80$ \\
\texttt{Hmm}              & $-36.00$ & $-0.61$  & $+35.41$ \\
\midrule
Prolong (Last) & $+0.45$  & $+0.76$  & $+0.31$ \\
Prolong (Mid)  & $+0.38$  & $+0.76$  & $+0.38$ \\
\bottomrule
\end{tabular}
\caption{
\textbf{Unconditional vs. error-targeted interventions} on GSM8K (Instruct model). Values report absolute accuracy change ($\%$) vs. baseline. Unconditional interventions often degrade performance, while predictor-gated interventions ($|\alpha|=0.05$) on only $12\%$ of examples yield additional gains.
}
\label{tab:intervention_summary}
\end{table}

\paragraph{Premise.} Test-time scaling methods modify generation by inserting tokens directly into the model's ongoing output stream. When the model is about to generate the final answer marker, we instead inject additional tokens that encourage further checking.\footnote{Specifically, we tested: \texttt{Wait} -- ``\texttt{Wait, let me double check.}'', \texttt{Hmm} -- ``\texttt{Hmm, let me think about this more carefully.}'', and \texttt{Check} -- ``\texttt{Let me double-check this step by step.}''} These injected cues encourage the model to extend its reasoning before committing to a final answer \citep{muennighoff2025s1simpletesttimescaling}.

Separately, activation steering intervenes on representations by adding previously derived activation directions that guide the model toward different behaviors \citep{zhang2024bestpracticesactivationpatching}. Rather than using cross-example activation differences, we construct \emph{in-sample} steering directions by averaging, within each prompt from the training split, the vector difference between \texttt{Step}-preceding and termination-preceding activations: 
\[
\mathbf{s}^{(\ell)} = \mathbb{E}_k\!\left[\mathbf{h}^{(\ell)}_{t(\text{term})-1} - \mathbf{h}^{(\ell)}_{t(\text{Step }k)-1}\right].
\]
During decoding on the test split, we intervene additively at the timestep immediately preceding final answer marker by updating the hidden state as $\mathbf{h}^{(\ell)}_t \leftarrow \mathbf{h}^{(\ell)}_t + \alpha\,\mathbf{s}^{(\ell)}$, where $\alpha$ controls the steering strength. We apply these interventions either at \textsc{Last}, steering the final five layers, or at \textsc{Mid}, steering the middle five layers. Subtracting the steering direction (\textsc{Prolong}; $\alpha < 0$) promotes extended reasoning, whereas adding it (\textsc{Shorten}; $\alpha > 0$) encourages earlier convergence toward termination. In this section, we focus on \textsc{Prolong} to induce additional reasoning and reflection.

\paragraph{Unconditional test-time scaling is often harmful on GSM8K.}
As summarized in Table~\ref{tab:intervention_summary}, always injecting control tokens into all examples reduces accuracy by as much as $11.7\%$ to $36.0\%$. Even comparatively style-native injection such as \texttt{Step} incurs a $1.59\%$ drop under unconditional use. These results provide further evidence that test-time scaling frequently perturbs already correct reasoning, introducing substantial collateral damage \citep{ghosal2025doesthinkinghelpmirage, wang2025waitdontneedwait}.

To mitigate these side effects, we move \textbf{toward error-targeted interventions that improve both accuracy and efficiency by selectively intervening only when failure is predicted.} Using correctness predictors from Section~\ref{sec:predictors}, we intervene on just $12.3\%$ of examples, converting unconditional accuracy drops into consistent gains of up to $+35.4\%$ relative to always-on interventions. 

\begin{figure*}[t!]
    \centering
    \includegraphics[width=\textwidth]{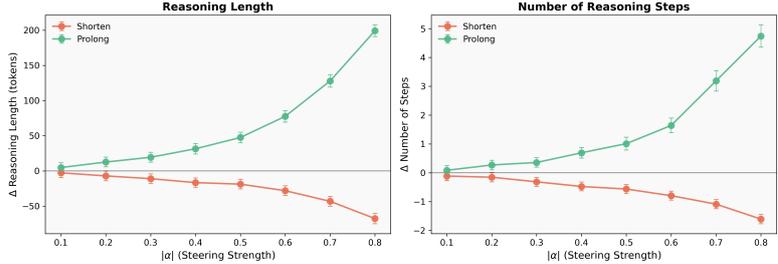}
    \caption{
    \textbf{Trajectory-based control of reasoning behavior.}
    (\textbf{a}) Correctness steering on GSM8K, stratified by original step count; values in $\%$.
    (\textbf{b}) Reasoning length control via the termination subspace as a function of steering strength~$|\alpha|$.
    }
    \label{fig:traj_based_control}
\end{figure*}

The modest net improvements observed across interventions reflect that \textbf{not all detectable errors are correctable via current inference-time interventions}. For example, the \texttt{Step}-injection intervention corrects only 26 of the 90 predictor-flagged incorrect reasoning instances. At the same time, the intervention reverts 14 originally correct solutions (flagged due to predictor imperfections) to incorrect. These opposing effects partially cancel, resulting in small net gains.

Taken together, these findings motivate error-targeted inference-time scaling: late-step trajectory geometry encodes detectable correctness signals that can be used to guide interventions, but fixed, one-shot corrections remain limited in their ability to reliably repair diverse failure modes.

\begin{table}[t]
\centering
\small
\setlength{\tabcolsep}{3.5pt}
\begin{tabular}{lcccccc}
\toprule
\textbf{Eval.\ Dataset} & \multicolumn{5}{c}{\textbf{Step}} & \textbf{Ans.} \\
 & 1 & 2 & 3 & 4 & 5 & \textbf{Marker} \\
\midrule
MATH-500 & 1.00 & 1.00 & 0.98 & 0.91 & 0.85 & 0.99 \\
MMLU     & 1.00 & 1.00 & 0.98 & 0.92 & 0.89 & 0.98 \\
Freeform GSM8K & 0.93 & 0.84 & 0.83 & 0.84 & 0.88 & 0.92 \\
\bottomrule
\end{tabular}
\caption{
\textbf{Cross-dataset and cross-format transfer of step probes from fixed-form GSM8K.}
Best-layer probe accuracy on held-out datasets. Structural geometry transfers robustly across tasks and formats.
}
\label{tab:cross_task_transfer}
\end{table}

\subsection{Cross-Task and Cross-Dataset Generalization}
\label{sec:cross_task}

To evaluate whether the trajectory framework generalizes beyond GSM8K, we extend our analyses to MATH-500 and the MMLU (validation set, 1,531 questions). Despite substantial differences in task difficulty or domain, GSM8K-trained step probes transfer strongly to both datasets.

\paragraph{Step-specific geometry transfers robustly across tasks and datasets.}
As shown in Table~\ref{tab:cross_task_transfer}, probes trained on GSM8K achieve best-layer accuracies above $0.85$ for all steps on both MATH-500 and MMLU, despite substantially different content: MATH-500 involves more complex mathematical problems, while MMLU spans diverse knowledge-intensive reasoning domains. Notably, although MMLU uses the same \texttt{Step X:} format as GSM8K, activations differ substantially between the two (cosine similarity $\approx 0.54$, CKA $\approx 0.60$). This further supports the interpretation from Section~\ref{sec:exp1} that step-specific representations capture reasoning progress rather than surface formatting cues. Beyond probe transfer, steering directions also generalize causally: \textsc{Prolong (Mid)} vectors derived from GSM8K improve MATH-500 accuracy from 36.40\% to 38.20\% (+1.80\%) without retuning, providing causal evidence that the identified geometry reflects general properties of mathematical reasoning rather than dataset-specific artifacts.

\paragraph{Correctness prediction is more task-sensitive.} A GSM8K-trained predictor achieves AUC $0.87$ in-distribution; when transferred, performance drops to $0.73$ on MATH-500, $0.64$ on MMLU, and $0.60$ on freeform GSM8K. This suggests a two-level organization: the \emph{structural} geometry of reasoning trajectories generalizes robustly across tasks, whereas \emph{correctness} geometry is shaped by task-specific error modes and solution-path diversity. Practically, the trajectory framework (step detection, boundary extraction, subspace identification) transfers without modification, while correctness predictors need further domain-specific calibration.

\section{Trajectory-Based Inference-Time Interventions}
\label{sec:exp3}

Motivated by the observations in Section~\ref{sec:error_aware_intervention}, we propose \emph{trajectory-based interventions}, which intervene adaptively based on how the reasoning trajectory evolves during generation. These interventions are effective both at correcting erroneous reasoning and at controlling reasoning length.

\subsection{Correcting Deviating Reasoning Trajectories}
Given the systematic divergence between correct and incorrect trajectories, we evaluate a low-rank steering strategy that guides the model back toward an ideal reasoning trajectory when the ongoing trajectory deviates beyond certain tolerance.

\paragraph{Methodology.}
From correct trajectories in the GSM8K training split, we extract activations immediately preceding each \texttt{Step} token and project them into a low-dimensional subspace via PCA. Let $z_j \in \mathbb{R}^d$ denote the projected activation at step $j$. The \emph{ideal trajectory} is the step-wise mean $\mu_j = \mathbb{E}[z_j]$ over correct examples, with a dispersion statistic $\sigma_j$ characterizing typical variability of correct reasoning at each step. This defines a step-indexed reference path with tolerance bands. 

Using held-out data, we optimize step-specific divergence thresholds that balance sensitivity to incorrect trajectories against false interventions on correct ones. At inference time, immediately before \texttt{Step}, activation is projected into the same subspace and compared against the ideal trajectory. We compute a local deviation $\delta_j = \lVert z_j - \mu_j \rVert$ and a cumulative deviation $D_j = \sum_{i<j} \delta_i$. If either $\delta_j$ or $D_j$ exceeds its corresponding threshold, we apply a low-rank steering update that moves the activation toward $\mu_j$ along the dominant principal directions. Because divergence is evaluated at every step, trajectory-based steering can intervene multiple times within a single example, correcting both abrupt deviations and gradual drift.  

\paragraph{Trajectory-based interventions are effective, and most effective on problems with longer reasoning chains.}
On GSM8K questions requiring six reasoning steps, accuracy improves from $75.44\%$ to $83.04\%$ ($+7.60\%$). Similarly, for seven-step problems, accuracy increases from $67.69\%$ to $75.38\%$ ($+7.69\%$). Notably, these gains are accompanied by high preservation rates ($\geq97\%$), indicating that repeated, low-magnitude interventions can correct difficult reasoning trajectories without destabilizing previously correct ones.

However, for problems with step count ${\leq}5$, the same method yields near-zero changes. This suggests that additional intervention provides limited benefit when the model already follows short, stable reasoning trajectories, consistent with Section~\ref{sec:exp2}, where early-step geometry exhibits minimal divergence between correct and incorrect solutions. 
Together, these results demonstrate that trajectory-based steering is most effective when applied to long, error-prone reasoning chains. By selectively nudging deviating trajectories back toward an ideal path, the method avoids the collateral damage and delivers robust net gains precisely where reasoning failures are most likely to occur.

\subsection{Reasoning Length Control}
We next use this termination-related subspace to \emph{directly and continuously control} reasoning length. We adopt the same steering directions and layer configurations as in Section~\ref{sec:exp2}, where each direction approximates the local trajectory toward the termination subspace. During decoding, we intervene additively to either push activations \emph{toward} this direction (\textsc{Shorten}), accelerating convergence and reducing reasoning length, or \emph{away from} it (\textsc{Prolong}), delaying convergence and extending intermediate computation. The intervention strength is controlled by a coefficient $|\alpha|$.

Figure~\ref{fig:traj_based_control} shows that for moderate steering strengths ($|\alpha| \leq 0.8$), reasoning length can be adjusted approximately monotonically with minimal impact on task accuracy (around 1\% change for $|\alpha| \leq 0.4$). This demonstrates that the termination-related subspace can act as a locally smooth control axis for reasoning length.
However, as $|\alpha|$ increases beyond this moderate range ($|\alpha| \gtrsim 0.8$), we observe behavioral mode collapse: the model enters repetitive loops in which step content is generated repeatedly without substantive progress. Therefore, increases in reasoning length for $|\alpha| \gtrsim 0.8$ do not reflect extended meaningful computation. Notably, this collapse is \emph{rare} at small steering strengths ($|\alpha| \leq 0.5$), where the loop ratio remains below $1\%$ and changes in length correspond to genuine extensions or shortening of reasoning.

\section{Discussion}
\subsection{Conclusion}

This work advances a geometric perspective on LLM reasoning by demonstrating that multi-step reasoning unfolds along structured trajectories in representation space. Intermediate reasoning states occupy step-specific regions that become increasingly linearly separable at deeper layers, and this organization is already present in Base models; reasoning distillation primarily reshapes the depth at which convergence occurs rather than introducing new representational structure.

Building on this, we show that late-step geometry provides an actionable signal for predicting final-answer correctness prior to answer emission, enabling more selective, error-targeted inference-time interventions that mitigate the overthinking associated with unconditional test-time scaling. Beyond correctness, reasoning trajectories can be causally manipulated to control reasoning length by steering activations toward or away from termination-related regions. These results advance the view of reasoning trajectories as a unifying abstraction for interpreting, predicting, and influencing LLM reasoning behavior.

\subsection{Future Work}
Our results suggest several directions for future work. 
First, trajectory-based analysis may offer a geometric perspective on CoT faithfulness: if a model's internal step activations diverge substantially from typical step-specific regions despite producing correct-sounding text, this could indicate unfaithful reasoning, complementing existing behavioral tests.
Second, a more fine-grained taxonomy of reasoning failures and the identification of error-specific geometric signatures could yield further insights.
Third, geometric insights could be incorporated directly into training objectives---for instance, step-level auxiliary losses that regularize hidden-state trajectories toward those observed in correct solutions, leveraging internal signals otherwise ignored by behavior-level supervision.
Finally, our analysis operates at the representation level. Bridging trajectory-level observations with circuit-level interpretability---identifying which attention heads and MLP neurons implement the observed dynamics---remains an important open direction.

\section*{Limitations}

This research has several limitations. First, although we observe clear and consistent trajectory structure in GSM8K, MATH-500, and MMLU, it remains an open question whether similar geometric organization arises in other settings, such as open-ended reasoning, multi-hop QA, or program synthesis.
Second, while we examine multiple training regimes (Base, Instruct, and reasoning-distilled models), our analysis is restricted to the Llama~3.1~8B family. Although the consistency of trajectory-level phenomena across three substantially different post-training objectives suggests that these structures are not paradigm-specific, we have not verified whether the same geometric organization holds at larger scales or across architecturally distinct model families. Larger models may exhibit qualitatively different trajectory dynamics due to increased capacity or different training distributions, and we consider cross-family and cross-scale exploration an important direction for future work.
Finally, our trajectory-based interventions rely on estimating an ideal trajectory from correct training examples. While this assumption is empirically supported in our settings, it may break down when correctness is underspecified. Extending this approach to such settings may require richer notions of ideal behavior, or task-conditioned reference trajectories. 
Still, revealing these trajectory-level structures offers a concrete lens for rethinking how reasoning unfolds, where errors arise, and how inference-time control might be made more precise.
This work serves as one step in that direction.

\section*{Ethical Considerations}
This work analyzes internal representations and inference-time interventions in LLMs using standard, publicly available benchmarks. It does not involve human subjects, personal data, or deployment in real-world decision-making contexts. Our study focuses on low-risk arithmetic reasoning tasks. We do not identify any additional ethical concerns beyond those generally associated with interpretability and model analysis research.

\newpage
\bibliography{custom}

\newpage

\appendix

\section{Fixed-form Prompt Setup}
\label{sec:prompt}

Across datasets, we use the following prompt to reliably elicit step-structured reasoning:
\begin{tcolorbox}[colback=gray!5,colframe=gray!50]
You are a helpful assistant that solves problems step by step with each step signified by ``Step [step\_number]: ''. Always provide your final answer after \texttt{\{final\_answer\_mark\}} at the end.

\textbf{Question:} \texttt{\{question\}}

Please solve this step by step, putting each step after ``Step [step\_number]: '' and always provide your final answer after \texttt{\{final\_answer\_mark\}}.

\textbf{Solution:}
\end{tcolorbox}

\section{Complete t-SNE Visualization}
\label{sec:t-sne}

\begin{figure*}[t]
    \centering

    \begin{subfigure}[t]{0.8\textwidth}
        \centering
        \includegraphics[width=0.98\textwidth]{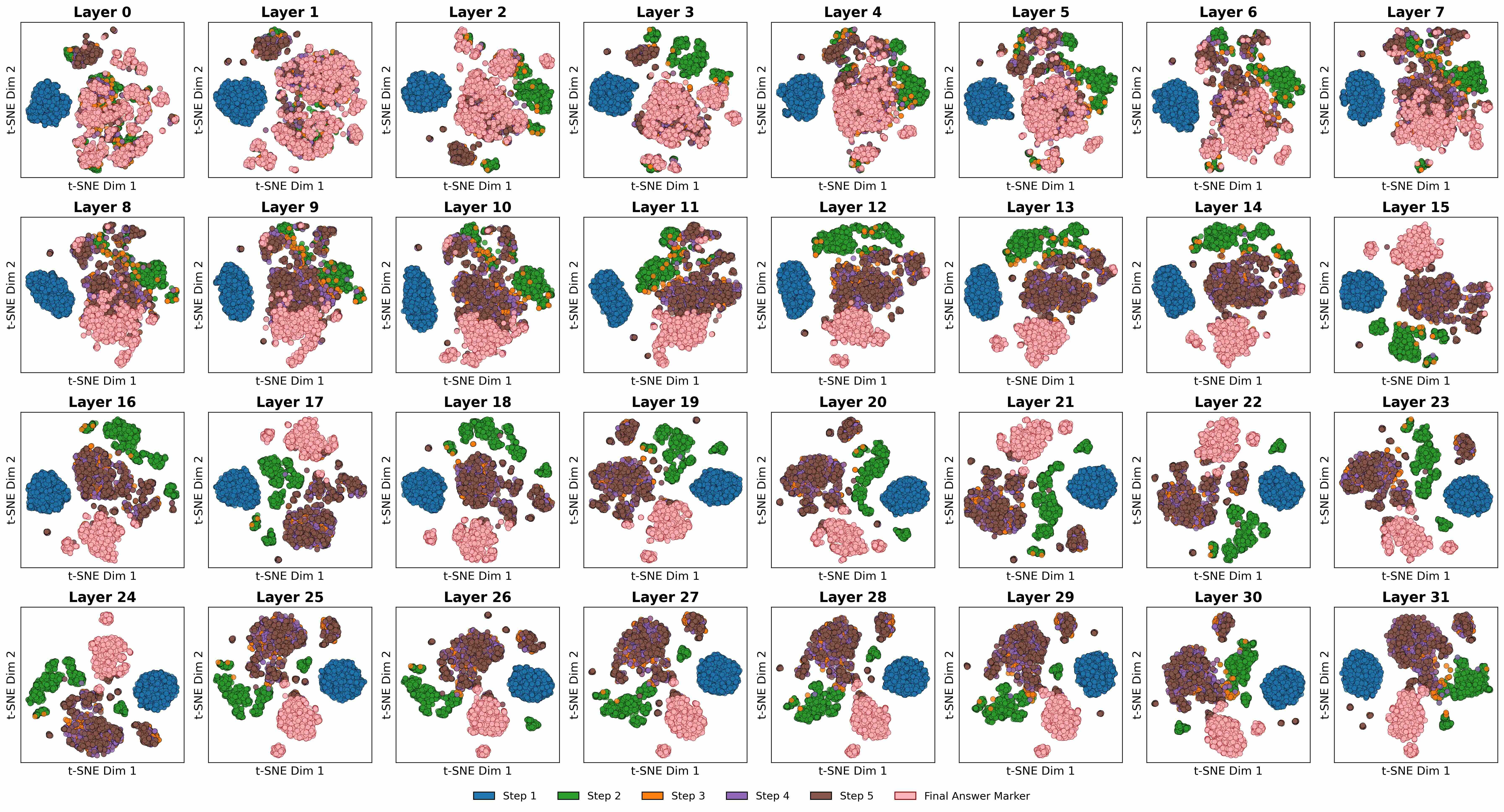}
        \caption{Instruct}
        \label{fig:appendix_tsne_inst}
    \end{subfigure}

    \vspace{0.6em}

    \begin{subfigure}[t]{0.8\textwidth}
        \centering
        \includegraphics[width=0.98\textwidth]{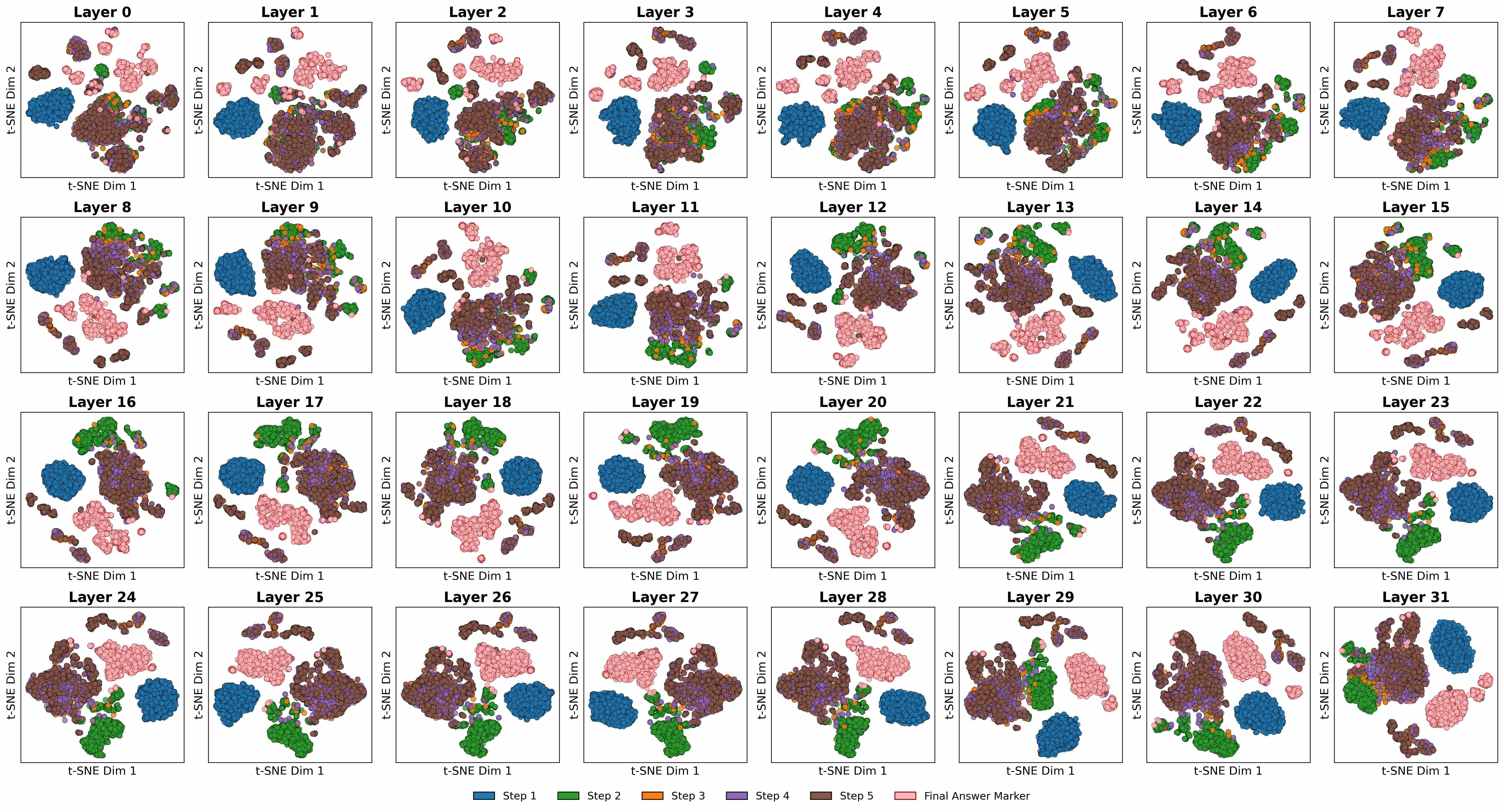}
        \caption{R1-Distilled}
        \label{fig:appendix_tsne_r1}
    \end{subfigure}

    \vspace{0.6em}

    \begin{subfigure}[t]{0.8\textwidth}
        \centering
        \includegraphics[width=0.98\textwidth]{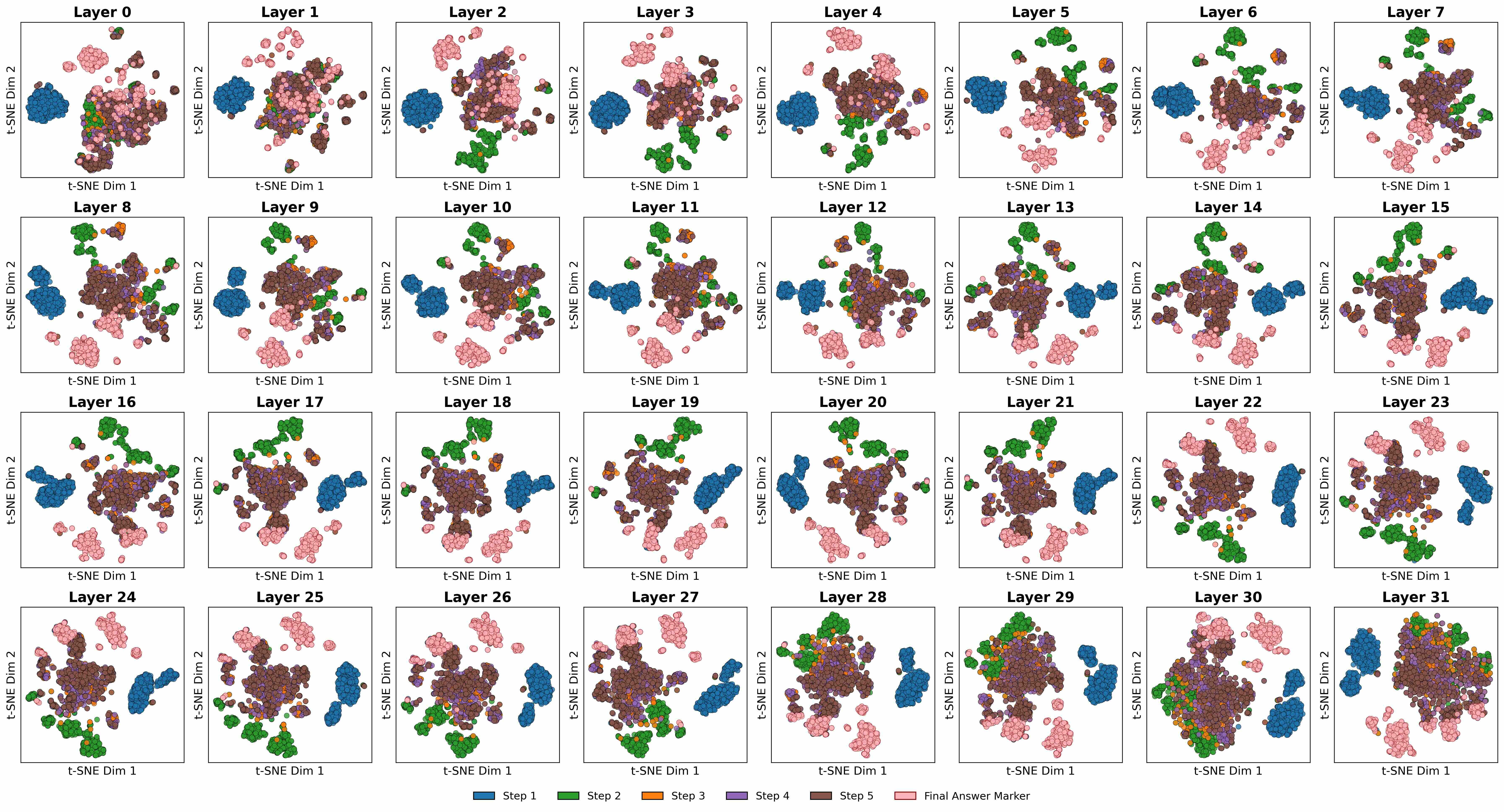}
        \caption{Base}
        \label{fig:appendix_tsne_base}
    \end{subfigure}

    \caption{Complete t-SNE visualizations of step-aligned hidden states extracted immediately before \texttt{Step} markers across three training regimes on the GSM8K test split. Each point corresponds to a residual-stream activation at a reasoning-step boundary. While all models exhibit step-structured organization, R1-Distilled and Instruct models show earlier and more pronounced separation between steps, whereas the Base model exhibits weaker but still discernible structure.}
    \label{fig:appendix_tsne}
\end{figure*}

See Figure~\ref{fig:appendix_tsne}.

\section{Complete Linear Probe Results}
\label{sec:linear_probe}
\begin{figure*}[h]
    \centering
    \includegraphics[width=\textwidth]{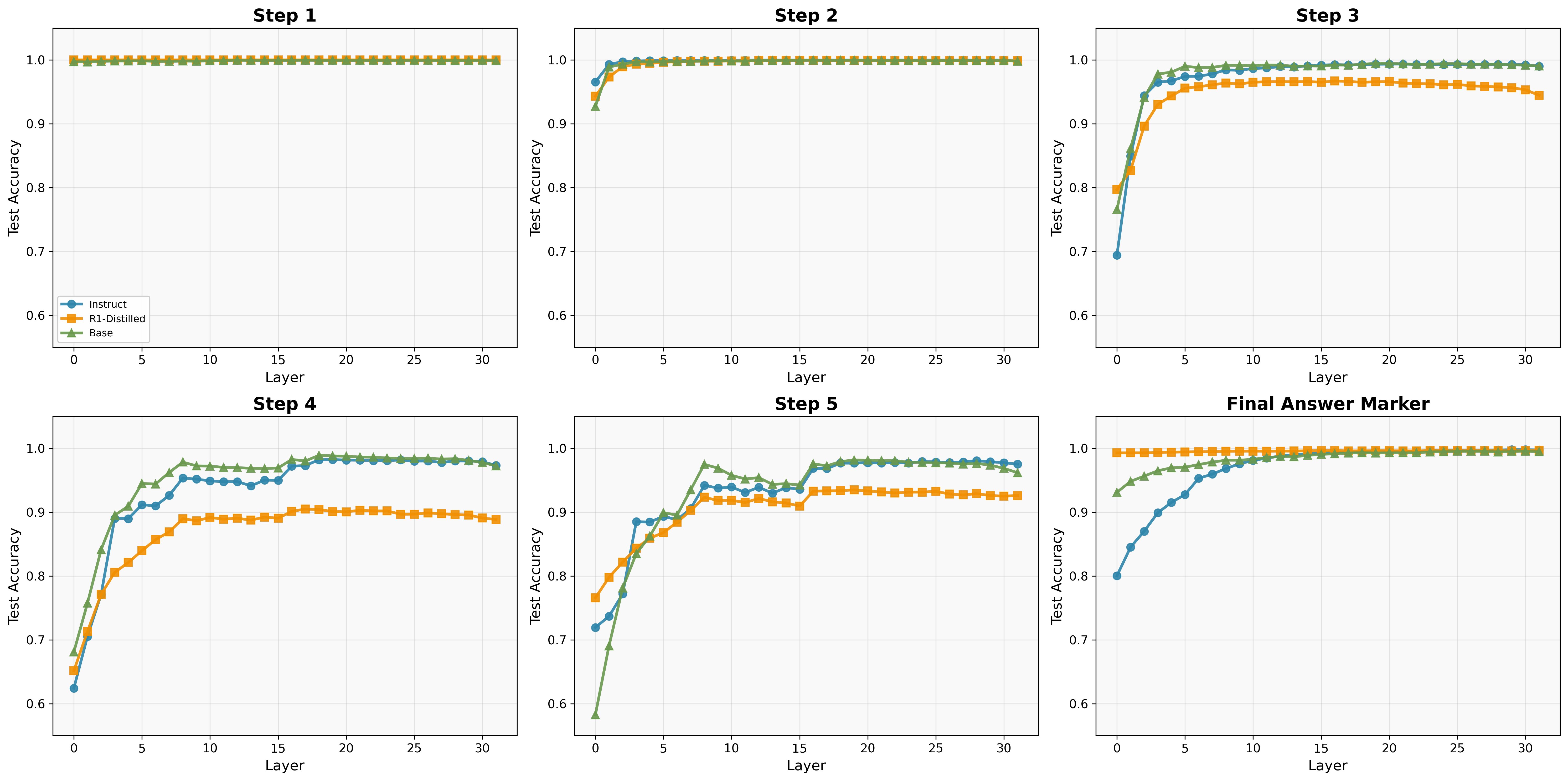}
    \caption{Layer-wise linear probe accuracy for predicting reasoning step identity. In each sub-figure, the $x$-axis denotes the layer from which activations are extracted, and the $y$-axis reports test accuracy of a linear classifier trained on these activations as input, with the reasoning step number (or the final answer marker) as the label. Early steps are linearly separable from shallow layers with near-ceiling accuracy, whereas later steps require deeper layers to become separable. Differences across training regimes indicate that instruction tuning and reasoning distillation shift the linear accessibility of step information.
    }
    \label{fig:appendix_linearprobe}
\end{figure*}

See Figure~\ref{fig:appendix_linearprobe} for complete within-model results; and Table~\ref{tab:linear_probe_transfer_avg} for averaged cross-model transfer results across layers. 

\begin{table*}[t]
\centering
\footnotesize
\setlength{\tabcolsep}{6pt}
\begin{tabular}{llcccccc}
\toprule
\textbf{Probe} &
\textbf{Eval.} &
\multicolumn{5}{c}{\textbf{Step}} &
\textbf{Final Ans.} \\
\textbf{From} &
\textbf{On} &
1 & 2 & 3 & 4 & 5 &
\textbf{Marker} \\
\midrule

Instruct
& Base     & 0.9976 & 0.9689 & 0.9182 & 0.8099 & 0.8785 & 0.8800 \\
& R1-Dist. & 0.9998 & 0.9753 & 0.8579 & 0.7705 & 0.8511 & 0.7826 \\
\midrule

R1-Dist.
& Base     & 0.9977 & 0.9381 & 0.8651 & 0.8344 & 0.8516 & 0.9102 \\
& Instruct & 1.0000 & 0.9057 & 0.7988 & 0.8065 & 0.8730 & 0.8819 \\
\midrule

Base
& Instruct & 1.0000 & 0.9900 & 0.9063 & 0.8714 & 0.8794 & 0.9197 \\
& R1-Dist. & 0.9936 & 0.9495 & 0.8237 & 0.8590 & 0.8883 & 0.9050 \\
\bottomrule
\end{tabular}

\caption{
\textbf{Average cross-model transfer accuracy of step-specific linear probes.}
Each entry reports the \emph{average} linear-probe accuracy across all layers when a classifier trained on step-specific activations from one model is evaluated on another model.
Results are averaged across transfer directions for each probe--evaluation pair.
Unlike Table~\ref{tab:linear_probe_transfer}, Step~1 is included, showing near-perfect transferability across models.
}
\label{tab:linear_probe_transfer_avg}
\end{table*}

\section{Trajectory Distance Difference}
\label{sec:appendix_traj_dist}
See Figure 6 below.
\begin{center}
\includegraphics[width=0.9\columnwidth]{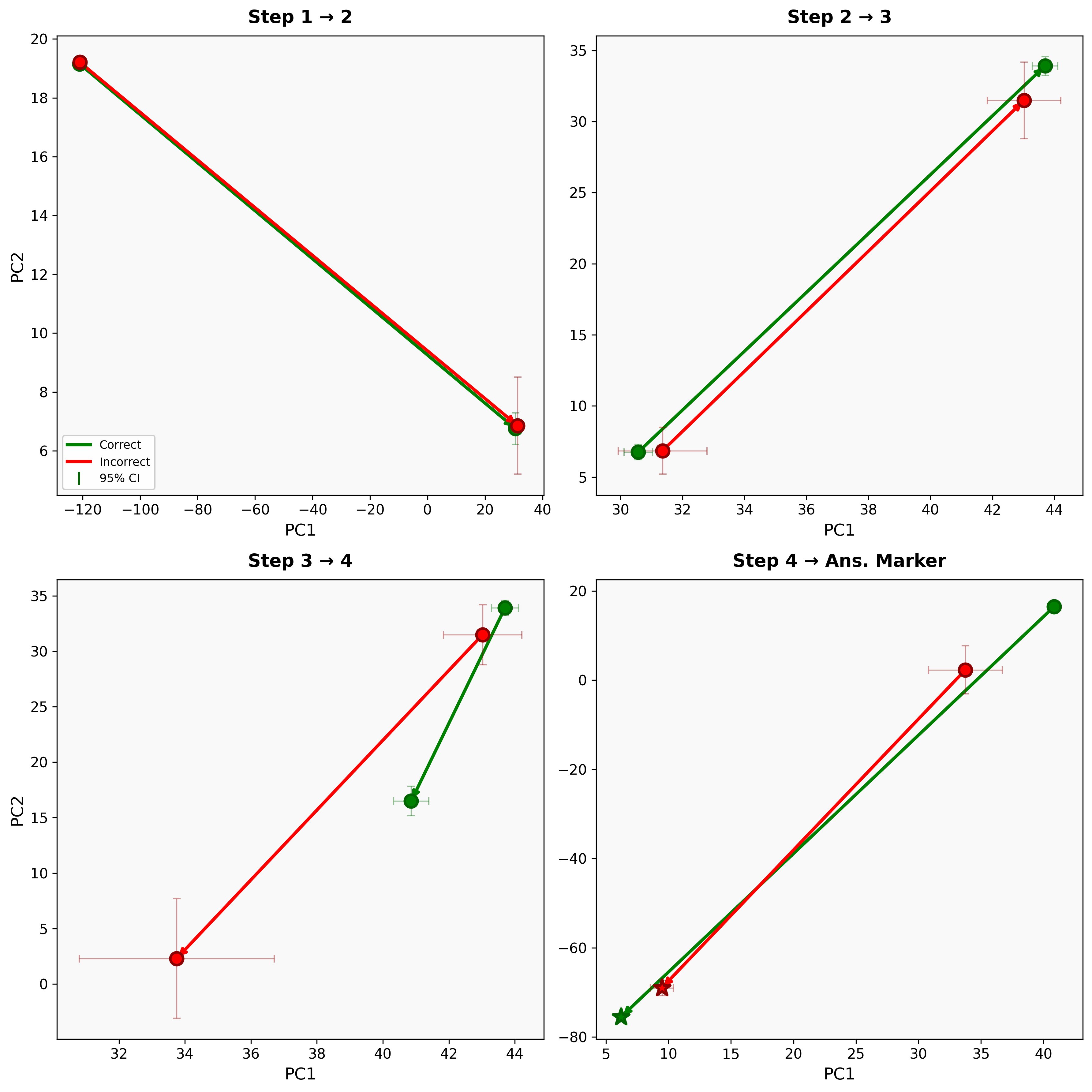}
\captionof{figure}{
\textbf{Between-step activation geometry differs between correct and incorrect reasoning.}
Results are from the Instruct model on the GSM8K train split with four reasoning steps; similar patterns hold across other step counts.
Late reasoning steps illustrate statistically significant geometric divergence for incorrect reasoning.
}
\label{fig:trajectory_segments}
\end{center}

\section{Freeform Generation Details}
\label{sec:freeform_details}

Table~\ref{tab:freeform_per_category} reports the full per-category breakdown of probe transfer from fixed-form to freeform GSM8K activations (Section~\ref{sec:freeform}). For each format category, the first number reports average probe accuracy across all 32 layers; the number in parentheses reports the best single-layer accuracy and the layer at which it is achieved. Probes were trained on fixed-form \texttt{Step X:} activations and evaluated without any retraining.

Across all format categories---including those with no \texttt{Step} markers in the surface text---probes achieve best-layer accuracies consistently above $0.82$, with most exceeding $0.85$. The ``Non-Step X:'' aggregate row pools all freeform examples that did \emph{not} spontaneously adopt \texttt{Step X:} formatting. Its strong performance ($\geq 0.84$ best-layer accuracy for all steps) confirms that the probes capture reasoning-progress information rather than surface-token identity.
\begin{table*}[t]
\centering
\footnotesize
\setlength{\tabcolsep}{4pt}
\begin{tabular}{lcccccc}
\toprule
\textbf{Category} & \textbf{Step 1} & \textbf{Step 2} & \textbf{Step 3} & \textbf{Step 4} & \textbf{Step 5} & \textbf{Answer} \\
\midrule
\texttt{Step X:}
  & 0.96 {\scriptsize(0.98 @ L1)} & 0.83 {\scriptsize(0.85 @ L19)} & 0.81 {\scriptsize(0.84 @ L28)} & 0.72 {\scriptsize(0.83 @ L10)} & 0.71 {\scriptsize(0.88 @ L30)} & 0.86 {\scriptsize(0.93 @ L20)} \\
Non-\texttt{Step X:}
  & 0.85 {\scriptsize(0.86 @ L1)} & 0.82 {\scriptsize(0.85 @ L18)} & 0.79 {\scriptsize(0.84 @ L7)} & 0.77 {\scriptsize(0.85 @ L10)} & 0.75 {\scriptsize(0.88 @ L25)} & 0.86 {\scriptsize(0.92 @ L10)} \\
\midrule
Numbered list
  & 0.82 {\scriptsize(0.91 @ L1)} & 0.83 {\scriptsize(0.89 @ L18)} & 0.81 {\scriptsize(0.83 @ L30)} & 0.75 {\scriptsize(0.85 @ L10)} & 0.73 {\scriptsize(0.89 @ L24)} & 0.86 {\scriptsize(0.95 @ L10)} \\
\texttt{\textbackslash n\textbackslash n} paragraphs
  & 0.82 {\scriptsize(0.82 @ L1)} & 0.75 {\scriptsize(0.81 @ L14)} & 0.79 {\scriptsize(0.83 @ L30)} & 0.82 {\scriptsize(0.86 @ L10)} & 0.81 {\scriptsize(0.91 @ L22)} & 0.81 {\scriptsize(0.96 @ L23)} \\
\texttt{\textbackslash n} lines
  & 0.87 {\scriptsize(0.87 @ L0)} & 0.82 {\scriptsize(0.87 @ L18)} & 0.77 {\scriptsize(0.86 @ L0)} & 0.75 {\scriptsize(0.86 @ L10)} & 0.79 {\scriptsize(0.90 @ L30)} & 0.89 {\scriptsize(0.94 @ L10)} \\
Single block
  & 0.87 {\scriptsize(0.88 @ L1)} & 0.87 {\scriptsize(0.87 @ L18)} & 0.81 {\scriptsize(0.87 @ L30)} & 0.74 {\scriptsize(0.85 @ L30)} & 0.67 {\scriptsize(0.85 @ L31)} & 0.87 {\scriptsize(0.94 @ L6)} \\
\bottomrule
\end{tabular}
\caption{
\textbf{Per-category freeform probe transfer accuracy.}
Each cell reports average accuracy across 32 layers, with the best single-layer accuracy and corresponding layer in parentheses. Probes trained on fixed-form \texttt{Step X:} activations transfer to all freeform format categories, including those with no \texttt{Step} markers.
}
\label{tab:freeform_per_category}
\end{table*}

\section{Experimental Details}
\label{sec:exp_details}
\paragraph{Generation.}
We use deterministic greedy decoding throughout (\texttt{do\_sample=False}).

\paragraph{Hidden-state extraction.}
We employ a two-pass procedure. In Pass~1, we run \texttt{model.generate()} with KV caching to produce the complete output sequence. In Pass~2, we concatenate the prompt and generated tokens and run a single forward pass with \texttt{output\_hidden\_states=True} and \texttt{use\_cache=False}. This yields 33 hidden-state vectors per token position (one from the embedding layer plus one from each of the 32 transformer layers), each of dimension $4{,}096$. These are the post-residual-add activations between transformer blocks---after both the self-attention and MLP sub-layers have been added to the residual stream, before the subsequent RMSNorm. We index position $t(\texttt{Step } k) - 1$ to obtain the activation immediately preceding each \texttt{Step} marker. Causal masking ensures that the single-pass hidden states are identical to those obtained during autoregressive generation.

\paragraph{Linear probes.}
We use logistic regression with \texttt{max\_iter=2000} and \texttt{class\_weight='balanced'}, retaining the library defaults for all other parameters (\texttt{solver='lbfgs'}, \texttt{penalty='l2'}, \texttt{C=1.0}). For each step label and each layer, we train a binary one-vs-rest classifier using an 80/20 stratified split.

\paragraph{Correctness predictors.}
We implement single-layer logistic regression (\texttt{nn.Linear($d$, 1)}) in PyTorch with binary cross-entropy loss and $\ell_2$ regularization via the Adam optimizer's \texttt{weight\_decay} parameter (set to $1/C$). Training uses learning rate $0.01$, batch size $32$, a maximum of $1{,}000$ epochs, and early stopping with patience $50$ on validation loss. The regularization strength $C$ is selected via 5-fold stratified cross-validation (\texttt{StratifiedKFold}) over $C \in \{0.001, 0.01, 0.1, 1.0, 10.0, 100.0\}$. Data is split 90/10 (stratified) for final training and evaluation. For feature sets requiring dimensionality reduction, PCA with $n_\text{components}=128$ is fit on the training split only.

\paragraph{Activation steering (Prolong/Shorten).}
Steering directions are computed per-layer as the mean difference between termination-preceding and step-preceding activations on the training split (Section~\ref{sec:error_aware_intervention}). During decoding, steering is applied additively at the token position immediately preceding the final answer marker. \textsc{Last} intervenes on the final 5 layers (layers 27--31); \textsc{Mid} intervenes on 5 layers centered at layer~15 (layers 13--17). The coefficient $|\alpha|$ controls the steering magnitude.

\paragraph{Trajectory-based steering.}
The ideal trajectory is estimated from correct-example activations at the final layer, projected via PCA ($n_\text{components}=128$, \texttt{random\_state=42}, fit on training split). The step-wise mean $\mu_j$ and dispersion $\sigma_j$ are computed over correct trajectories in PCA space. Low-rank steering updates use rank $r=32$: the correction is $\alpha \cdot \Delta z_j^{(r)} U_r$, where $U_r \in \mathbb{R}^{r \times d}$ contains the top-$r$ principal components and $\Delta z_j^{(r)}$ is the displacement toward $\mu_j$ in the $r$-dimensional subspace. Per-step divergence thresholds are optimized on held-out data to balance detection sensitivity against false interventions.

\paragraph{Seed-variance analysis.}
All generation is deterministic (greedy decoding), so variance arises only from classifier training. We report results over seeds $\in \{42, 123, 456\}$ in Table~\ref{tab:seed_variance}.

\begin{table*}[h]
\centering
\small
\setlength{\tabcolsep}{6pt}
\begin{tabular}{lc}
\toprule
\textbf{Component} & \textbf{Result} \\
\midrule
Linear probes (best-layer acc.) & $\leq \pm 0.001$ across all steps \\
Correctness pred.\ (Final State) & $0.837 \pm 0.016$ (L29) \\
Correctness pred.\ (Final State PCA) & $0.859 \pm 0.023$ (L19) \\
Correctness pred.\ (Late Steps PCA) & $0.853 \pm 0.038$ (L21) \\
\bottomrule
\end{tabular}
\caption{
\textbf{Seed-variance analysis} for key components. Linear probe accuracies are highly stable; correctness predictor AUCs show modest variance consistent with the smaller effective sample size of incorrect examples.
}
\label{tab:seed_variance}
\end{table*}

\section{Computational Overhead of Trajectory Steering}
\label{sec:overhead}

Trajectory-based steering operates only at step boundaries (typically 3--7 per problem), not at every generated token. Table~\ref{tab:overhead} summarizes the per-problem steering cost for rank $r{=}32$.

\begin{table*}[h]
\centering
\small
\setlength{\tabcolsep}{6pt}
\begin{tabular}{lcc}
\toprule
\textbf{Operation} & \textbf{Per-call} & \textbf{Total} \\
\midrule
PCA projection + divergence check & ${\sim}12.9\,\mu$s & ${\sim}81\,\mu$s \\
Low-rank update ($r{=}32$) & ${\sim}10.8\,\mu$s & ${\sim}24\,\mu$s \\
\midrule
Trajectory steering (combined) & --- & ${\sim}105\,\mu$s \\
Prolong (per-token $\times$ all timesteps) & ${\sim}1.12\,\mu$s & ${\sim}3{,}371\,\mu$s \\
\bottomrule
\end{tabular}
\caption{
\textbf{Steering computation cost per problem} on a single NVIDIA H200 GPU with Llama-3.1-8B-Instruct. Trajectory steering at step boundaries is approximately $32\times$ cheaper than per-token vector addition. Totals assume ${\sim}7$ step boundaries and ${\sim}3{,}000$ generated tokens, respectively.
}
\label{tab:overhead}
\end{table*}

The observed $1.38\times$ end-to-end wall-clock slowdown ($6.69$s $\rightarrow$ $9.21$s per question) does not arise from the steering operations themselves. In our research prototype, steered generation uses a manual token-by-token loop with \texttt{output\_hidden\_states=True}, which disables KV caching; the full growing sequence is therefore reprocessed at every token. An optimized implementation preserving KV caching would eliminate this overhead, as the algorithmic cost of trajectory steering is negligible relative to the forward pass.

\section{Step-Count Distributions Across Models}
\label{sec:step_count_dist}

Table~\ref{tab:step_count_dist} reports the distribution of reasoning step counts across the three model variants on GSM8K. Step counts differ substantially: Instruct produces shorter reasoning chains (median $K{=}4$) concentrated in 3--5 steps, while Base and R1-Distill produce longer chains (median $K{=}6$ and $K{=}5$, respectively). Despite these differences, cross-model probe transfer accuracy consistently exceeds $0.90$ (Table~\ref{tab:linear_probe_transfer}), indicating that the step-indexed geometry reflects a consistent representational subspace for reasoning progress that is robust to absolute position in the chain. Within-model trajectory analyses (Section~\ref{sec:predictors}) rely on last-step and step-transition features defined relative to each example's own trajectory length, so step-count differences do not confound these results.

\begin{table}[h]
\centering
\small
\setlength{\tabcolsep}{6pt}
\begin{tabular}{lccc}
\toprule
 & \textbf{Instruct} & \textbf{Base} & \textbf{R1-Distill} \\
\midrule
Mean $K$   & 4.5 & 6.6 & 6.9 \\
Median $K$ & 4   & 6   & 5   \\
$K \in [3,5]$ & 70.5\% & 33.4\% & 52.1\% \\
\bottomrule
\end{tabular}
\caption{
\textbf{Reasoning step-count distributions on GSM8K} across training regimes. Instruct produces substantially shorter chains than Base and R1-Distill, yet step-specific probes transfer across models with high accuracy, indicating that the geometry is tied to reasoning progress rather than absolute step position.
}
\label{tab:step_count_dist}
\end{table}

\end{document}